\documentclass[twoside,11pt]{article}

\usepackage{jmlr2e}
\usepackage{booktabs}
\usepackage{multirow}
\usepackage{array}
\usepackage{xcolor}
\usepackage{colortbl}
\usepackage{enumitem}
\usepackage{adjustbox}
\usepackage{lscape}
\usepackage{listings}
\usepackage{url}
\usepackage{amsmath}
\usepackage{amssymb}
\usepackage{pifont}
\usepackage{subcaption}

\newcommand{\cmark}{\textcolor{green!70!black}{\ding{51}}}
\newcommand{\xmark}{\textcolor{gray!60}{---}}
\newcommand{\lib}{{\textsf{LIDARLearn}}}

\ShortHeadings{LIDARLearn: A Unified Library for 3D Point Cloud Deep Learning}{Ohamouddou et al.}
\firstpageno{1}

\begin{document}
	
	\title{\lib{}: A Unified Deep Learning Library for \\
		3D Point Cloud Classification, Segmentation, \\
		and Self-Supervised Representation Learning}
	
	\author{\name Said Ohamouddou \email said\_ohamouddou@um5.ac.ma \\
		\addr ENSIAS, Mohammed V University, Rabat, Morocco \\
		\AND
		\name Hanaa El Afia \\
		\addr ENSIAS, Mohammed V University, Rabat, Morocco \\
		\AND
		\name Abdellatif El Afia \\
		\addr ENSIAS, Mohammed V University, Rabat, Morocco \\
		\AND
		\name Raddouane Chiheb \\
		\addr ENSIAS, Mohammed V University, Rabat, Morocco}
	
	\editor{TBD}
	
	\maketitle
	
	\begin{abstract}
		Three-dimensional (3D) point cloud analysis has become central to applications ranging from autonomous driving and robotics to forestry and ecological monitoring.
		Although numerous deep learning methods have been proposed for point cloud understanding, including supervised backbones, self-supervised pre-training (SSL), and parameter-efficient fine-tuning (PEFT), their implementations are scattered across incompatible codebases with differing data pipelines, evaluation protocols, and configuration formats, making fair comparisons difficult.
		We introduce \lib{}, a unified, extensible PyTorch library that integrates over 55 model configurations covering 29 supervised architectures, seven SSL pre-training methods, and five PEFT strategies, all within a single registry-based framework supporting classification, semantic segmentation, part segmentation, and few-shot learning.
		\lib{} provides standardised training runners, cross-validation with stratified $K$-fold splitting, automated LaTeX/CSV table generation, built-in Friedman/Nemenyi statistical testing with critical-difference diagrams for rigorous multi-model comparison, and a comprehensive test suite with 2\,200+ automated tests validating every configuration end-to-end.
		The code is available at \url{https://github.com/said-ohamouddou/LIDARLearn} under the MIT licence.
	\end{abstract}
	
	\begin{keywords}
		Point Clouds, 3D Deep Learning, Self-Supervised Learning, Parameter-Efficient Fine-Tuning, Open Source Software
	\end{keywords}
	
	\section{Introduction}
	\label{sec:intro}
	
	Light Detection and Ranging (LiDAR) technology has become a cornerstone of 3D scene understanding across scientific and engineering disciplines.
	In forestry, airborne and terrestrial laser scanning generate complex point cloud datasets that require advanced processing to extract structural metrics, such as canopy height and above-ground biomass~\cite{Fassnacht2016}.
	In autonomous driving and indoor robotics, real-time semantic segmentation of dense point clouds is critical for navigation and scene understanding~\cite{wu2024point}.
	The emergence of deep learning methods that operate directly on unstructured point sets, beginning with PointNet~\cite{qi2017pointnet} and PointNet++~\cite{qi2017pointnet++}, has driven rapid progress, spawning dozens of architectures that exploit local geometry~\cite{wang2019dynamic,curvenet}, attention mechanisms~\cite{guo2021pct,zhao2021point,wu2022point,wu2024point}, self-supervised pre-training~\cite{pang2022masked,yu2022point,chen2024pointgpt,dong2023act,qi2023contrast}, and parameter-efficient transfer~\cite{zha2023instance,zhou2024dynamic,sun24ppt,PointGST}.
	
	Despite this progress, the practical impact is limited by fragmentation; each method ships in its own repository with bespoke data loaders, evaluation scripts, and configuration formats.
	Reproducing or extending published results requires significant engineering efforts to reconcile these differences.
	Existing open-source libraries partially address this---OpenPoints~\cite{qian2022pointnext} supports ${\sim}15$ classification backbones, Pointcept~\cite{wu2023maskedscenecontrast} focuses on segmentation, and Torch-Points3D~\cite{torchpoints3d} provides a modular framework---but none unifies supervised training, SSL pre-training, and PEFT fine-tuning across all three task types (classification, semantic segmentation, and part segmentation) under a single consistent interface.
	
	Here, we introduce \lib{}, a comprehensive library that fills this gap.
	\lib{} offers: (i)~29 supervised classification/segmentation backbones and 7 SSL pre-training methods faithfully ported from their original repositories and verified by dual code audits; (ii)~5 PEFT strategies applicable to 4 SSL backbones, yielding 20 additional fine-tuning configurations; (iii)~standardised training pipelines supporting classification, semantic segmentation, part segmentation, and few-shot learning; (iv)~rigorous evaluation infrastructure including stratified $K$-fold cross-validation, per-class metrics via \texttt{scikit-learn}, and automated LaTeX table generation; and (v)~a 2\,200-test automated test suite covering configuration parsing, forward-pass shape verification, pre-trained checkpoint loading, and script coverage.
	We demonstrate the library by benchmarking all 56 configurations on a LiDAR tree species classification task (Appendix~\ref{app:application}).

	\section{Advantages and Comparison to Existing Toolkits}
	\label{sec:comparison}
	
	Table~\ref{tab:comparison} compares \lib{} with five established point cloud libraries across key capability dimensions.
	
	\begin{table}[t]
		\centering
		\caption{Comparison of \lib{} with existing point cloud deep learning libraries.
			\emph{Cls} = classification models, \emph{Seg} = semantic segmentation models,
			\emph{Part} = part segmentation models.}
		\label{tab:comparison}
		\renewcommand{\arraystretch}{1.15}
		\small
		\begin{adjustbox}{center}
			\begin{tabular}{lcccccc}
				\toprule
				\textbf{Feature} & \textbf{Learning3D} & \textbf{Open3D-ML} & \textbf{OpenPoints} & \textbf{Pointcept} & \textbf{Torch-Points3D} & \textbf{\lib{}} \\
				\midrule
				Cls models       & 5  & 0  & 15 & 3  & 6  & \textbf{29} \\
				Seg models       & 2  & 5  & 9  & 11 & 10 & \textbf{26} \\
				Part-seg models  & 0  & 0  & 5  & 1  & 2  & \textbf{26} \\
				SSL pre-training & 0  & 0  & 1  & 1  & 0  & \textbf{7}  \\
				PEFT strategies  & 0  & 0  & 0  & 0  & 0  & \textbf{5}  \\
				Few-shot eval    & \xmark & \xmark & \xmark & \xmark & \xmark & \cmark \\
				$K$-fold CV      & \xmark & \xmark & \xmark & \xmark & \xmark & \cmark \\
				Automated tests  & \xmark & \xmark & \xmark & \xmark & \xmark & \cmark \\
				\bottomrule
			\end{tabular}
		\end{adjustbox}
	\end{table}
	
	\textbf{Model Diversity.}
	\lib{} integrates 29~supervised backbones, 7~SSL pre-training methods, and 5~PEFT strategies.
	The supervised backbone spans three families.
	\emph{Point-based methods:}
	DELA~\cite{dela}, PPFNet~\cite{ppfnet}, PointCNN~\cite{pointcnn}, PointConv~\cite{pointconv}, PointKAN~\cite{pointkan}, PointMLP~\cite{pointmlp}, PointNet~\cite{qi2017pointnet}, PointNet++~\cite{qi2017pointnet++}, PointSCNet~\cite{pointscnet}, PointWeb~\cite{pointweb}, RSCNN~\cite{rscnn}, RepSurf~\cite{repsurf}, and SONet~\cite{sonet}.
	\emph{Attention and transformer architectures:}
	GlobalTransformer~\cite{pointtnt}, P2P~\cite{p2p}, PCT~\cite{guo2021pct}, PVT~\cite{pvt}, PointTNT~\cite{pointtnt}, PointTransformer~\cite{zhao2021point}, PointTransformerV2~\cite{wu2022point}, and PointTransformerV3~\cite{wu2024point}.
	\emph{Graph neural networks:}
	CurveNet~\cite{curvenet}, DGCNN~\cite{wang2019dynamic}, DeepGCN~\cite{deepgcn}, GDAN~\cite{gdanet}, KANDGCNN~\cite{kandgcnn}, MSDGCNN~\cite{msdgcnn}, and MSDGCNN2~\cite{msdgcnn2}.
	For \emph{self-supervised pre-training}, seven methods are included:
	ACT~\cite{dong2023act}, PCP-MAE~\cite{pcpmae}, Point-BERT~\cite{yu2022point}, PointGPT~\cite{chen2024pointgpt}, Point-M2AE~\cite{zhang2022point}, Point-MAE~\cite{pang2022masked}, and ReCon~\cite{qi2023contrast}.
	Five \emph{parameter-efficient fine-tuning} strategies are supported:
	DAPT~\cite{zhou2024dynamic}, IDPT~\cite{zha2023instance}, PPT~\cite{sun24ppt}, PointGST~\cite{PointGST}, and VPT~\cite{jia2022visual} were integrated with four SSL backbones (PointMAE, PointGPT, ACT, and ReCon), producing 20 additional fine-tuning configurations.
	
	\textbf{SSL Faithfulness.}
	Each SSL implementation was audited against its original repository in two independent rounds, verifying the core model logic, loss computation, checkpoint prefix-stripping, freezing behavior, and classification/segmentation output consistency.
	All critical issues found during the audits were fixed and re-verified (see Appendix~\ref{app:audit}).
	
	\textbf{Evaluation Rigour.}
	\lib{} implements stratified $K$-fold cross-validation with per-fold model rebuild, optimizer reset, and deterministic seeding---preventing state leakage between folds.
	Metrics (accuracy, balanced accuracy, macro/weighted F1, Cohen's $\kappa$, per-class precision/recall) were computed via \texttt{scikit-learn} at the best-validation-accuracy epoch.
	Few-shot evaluation follows the standard 10-episode protocol on ModelNet40 with automated mean$\pm$std aggregation.
	The results were exported as LaTeX tables with per-column best-value highlighting and CSV files for programmatic analysis.

	\section{Toolbox Usage}
	\label{sec:usage}
	
	\subsection{Dependencies and Installation}
	
	\lib{} requires Python~$\geq$3.8, PyTorch~$\geq$1.10, and \texttt{timm}.
	The required CUDA extensions (\texttt{chamfer\_dist}, \texttt{pointnet2\_ops}, \texttt{dela\_cutils}) are built via \texttt{scripts/install\_extensions.sh}.
	The library uses a YAML-based configuration system with \texttt{\_base\_} inheritance, where the dataset parameters, model hyperparameters, and optimizer settings are cleanly separated.
	
	\subsection{Workflow}
	
	The following example trains PointMAE with DAPT fine-tuning on a custom LiDAR tree species dataset using 5-fold cross-validation.
	
	\begin{lstlisting}
		# 1. Pre-train on ShapeNet55 (300 epochs)
		python main.py --config cfgs/classification/PointMAE/ShapeNet55/shapenet55_pretrain.yaml \
		--mode pretrain --exp_name mae_pretrain
		
		# 2. Fine-tune with DAPT on STPCTLS (5-fold CV)
		python main.py --config cfgs/classification/PointMAE/STPCTLS/stpctls_cv_dapt.yaml \
		--ckpts pretrained/pretrained_mae.pth --run_all_folds --seed 42
		
		# 3. Generate LaTeX comparison table
		python scripts/generate_comparison_table_one_table.py \
		--exp_dir experiments/STPCTLS
	\end{lstlisting}
	
	All 56 classification model/strategy configurations can be launched using a single command: \texttt{bash scripts/train\_stpctls.sh}.
	Equivalent sweep scripts exist for ModelNet40, S3DIS segmentation, ShapeNetParts part segmentation, and ModelNet40 few-shot evaluation.
	For rigorous model comparison across $k$-fold cross-validation runs, \lib{} also provides \texttt{scripts/friedman\_cv\_analysis.py}, which performs the Friedman non-parametric test with Nemenyi post-hoc analysis and critical-difference diagrams over the per-fold scores of all models, producing a publication-ready LaTeX output.
	
	\subsection{Architecture}
	
\begin{figure}[t]
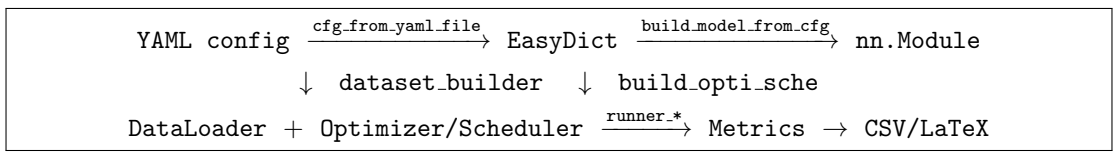

		\centering
		\fbox{\parbox{0.95\textwidth}{\centering
				\small
				\texttt{YAML config} $\;\xrightarrow{\texttt{cfg\_from\_yaml\_file}}\;$
				\texttt{EasyDict} $\;\xrightarrow{\texttt{build\_model\_from\_cfg}}\;$
				\texttt{nn.Module}\\[4pt]
				$\downarrow$\quad \texttt{dataset\_builder} \quad$\downarrow$\quad \texttt{build\_opti\_sche}\\[4pt]
				\texttt{DataLoader} $\;+\;$ \texttt{Optimizer/Scheduler}
				$\;\xrightarrow{\texttt{runner\_*}}\;$
				\texttt{Metrics} $\;\to\;$ \texttt{CSV/LaTeX}
		}}
		\caption{Simplified \lib{} pipeline. YAML configs drive model construction, dataset loading, and optimizer building via a shared registry. Task-specific runners produce standardised metric outputs.}
		\label{fig:arch}
	\end{figure}

	Figure~\ref{fig:arch} illustrates this pipeline.
	Models are registered via \texttt{@MODELS.register\_module()} decorators and built from YAML configs through a unified \texttt{build\_model\_from\_cfg} factory.
	Five task-specific runners (\texttt{runner\_finetune}, \texttt{runner\_pretrain}, \texttt{runner\_seg}, etc.) share a common infrastructure for checkpointing, metrics tracking, and logging, whereas each implements task-specific forward/loss logic.
	The PEFT optimizer builder (\texttt{\_build\_peft\_optimizer}) automatically freezes the backbone parameters and trains only strategy-specific adapters/prompts based on a single \texttt{optimizer.part} config field.
	
	\subsection{Adding a New Model}
	
	Integrating a new classification backbone requires three steps.
	
	\begin{enumerate}[nosep,leftmargin=*]
		\item \textbf{Model file.} Create \texttt{models/mymodel/mymodel.py} with a class
		that inherits from \texttt{BasePointCloudModel} and is decorated with
		\texttt{@MODELS.register\_module()}. The class must implement
		\texttt{forward(pts)} returning \texttt{[B, num\_classes]} logits;
		\texttt{get\_loss\_acc} is inherited automatically.
		
		\item \textbf{Registration.} Add \texttt{import models.mymodel.mymodel} to
		\texttt{models/\_\_init\_\_.py} so the registry discovers the class at
		import time.
		
		\item \textbf{Config.} A YAML file (for example,\
		\texttt{cfgs/classification/MyModel/STPCTLS/stpctls.yaml}) specifying
		\texttt{model.NAME: MyModel} alongside dataset, optimizer, and scheduler
		fields. The model is now trainable via
		\texttt{python main.py --config <yaml> --mode finetune}.
	\end{enumerate}
	
	\noindent
	Segmentation models follow the same pattern but inherit from
	\texttt{BaseSegModel} and implement \texttt{forward(pts, cls\_label)}
	returning the per-point logits. No changes to the runners, metrics, or
	testing infrastructure are needed---the new model is automatically
	covered by the L1 config parse test suite.
	
	\section{Conclusion}
	\label{sec:conclusion}
	
	We presented \lib{}, an open-source library that unifies 29 supervised backbones, seven SSL pre-training methods, and five PEFT strategies for 3D point cloud analysis under a single, tested, and extensible framework.
	\lib{} supports classification, semantic segmentation, part segmentation, and few-shot evaluation with rigorous cross-validation and automated reporting.
	With 2\,200+ automated tests and dual-round SSL implementation audits, the library is designed for reproducible research and fair benchmarking.
	As a case study, we benchmarked all 56 model configurations on the STPCTLS terrestrial laser scanning dataset for LiDAR tree species classification (Appendix~\ref{app:application}), demonstrating the library's ability to conduct comprehensive and standardized comparisons across supervised, SSL, and PEFT methods for a real-world domain-specific task.

	
	\bibliography{references,comparison_with_other_lbararie,why_Lidar_data_processing.tex}
	
	
	\newpage
	\appendix
	
	\section{Supported Models}
	\label{app:models}
	\emph{The complete 56-configuration support matrix is provided in the online repository.}
	
	\section{SSL Audit Summary}
	\label{app:audit}
	
	Each of the seven SSL methods was audited in two independent rounds against its original GitHub repository.
	The audit verified (i) the core encoder/decoder architecture and loss functions.
	(ii)checkpoint loading with correct prefix-stripping;
	(iii) Parameter freezing behavior during fine-tuning
	(iv) Classification and segmentation output consistency.
	All 7~SSL methods were verified as being faithful to their original implementations after the fixes were applied.
	
	\section{Application: LiDAR Tree Species Classification}
	\label{app:application}
	
	As a case study, we evaluated all 56 model configurations on the STPCTLS terrestrial laser scanning dataset~\cite{seidelPredictingTreeSpecies2021c}, which contains high-resolution point clouds from artificial and natural forests in Germany and the United States.
	Seven tree species---beech, red oak, ash, oak, douglas fir, spruce, and pine---were selected for their significant morphological similarity across species and considerable within-species variation owing to diverse growth conditions, making the classification task challenging.
	To further increase the difficulty, we added isotropic Gaussian noise ($\sigma=0.20$) to the normalized point clouds.
	The dataset was divided into training and testing sets using an 80:20 ratio.
	The sample sizes and class distributions are presented in Table~\ref{tab:dataset_summary}.
	
	\begin{table}[h!]
		\centering
		\scriptsize
		\caption{Train and test distribution for the STPCTLS dataset.}
		\label{tab:dataset_summary}
		\renewcommand{\arraystretch}{1.1}
		\begin{tabular}{lcccr}
			\toprule
			\textbf{Species} & \textbf{Total} & \textbf{Train} & \textbf{Test} & \textbf{\%} \\
			\midrule
			Douglas Fir (\textit{Pseudotsuga menziesii}) & 183 & 146 & 37 & 26.5 \\
			Beech (\textit{Fagus sylvatica}) & 164 & 131 & 33 & 23.7 \\
			Spruce (\textit{Picea abies}) & 158 & 126 & 32 & 22.9 \\
			Red Oak (\textit{Quercus rubra}) & 100 & 80 & 20 & 14.5 \\
			Ash (\textit{Fraxinus excelsior}) & 39 & 31 & 8 & 5.6 \\
			Pine (\textit{Pinus sylvestris}) & 25 & 20 & 5 & 3.6 \\
			Oak (\textit{Quercus robur}) & 23 & 18 & 4 & 3.2 \\
			\midrule
			\textbf{Total} & \textbf{692} & \textbf{552} & \textbf{139} & \textbf{100.0} \\
			\bottomrule
		\end{tabular}
	\end{table}
	
	Tables~\ref{tab:cv_comparison} and~\ref{tab:cv_perclass} are auto-generated from the cross-validation summaries via \\ \texttt{scripts/generate\_comparison\_table\_one\_table.py}.
	Table~\ref{tab:cv_comparison} presents classification accuracy, balanced accuracy, macro F1, recall, and precision for all 56 model configurations, grouped by category.
	Among the supervised methods, PointSCNet~\cite{pointscnet} and PointNet++ (MSG)~\cite{qi2017pointnet++} achieve the best performance.
	Among the SSL backbones, ACT with PointGST fine-tuning achieved the highest overall accuracy (69.78
	The per-class recall (Table~\ref{tab:cv_perclass}) reveals that minority species (Ash, Oak, and Pine) remain challenging for all methods, with the best per-class recall on Ash reaching only 50
	
	\begin{table}[htbp]
		\centering
		\scriptsize
		\caption{STPCTLS classification results: performance metrics and model parameters.}
		\label{tab:cv_comparison}
		\begin{adjustbox}{center}
			\begin{tabular}{p{1.8cm}p{2.3cm}p{1cm}p{1.5cm}p{1cm}p{1cm}p{1cm}p{1cm}p{1cm}p{1cm}p{1cm}}
				\toprule
				Category & Model & Strategy & Accuracy (\%) & Balanced Acc (\%) & F1 Macro (\%) & Recall (\%) & Precision (\%) & Total Params (M) & Train Params (M) & Epoch Time (s) \\
				\midrule
				\multirow{15}{*}{Point-based} & PointNet & - & 51.08 & 46.02 & 43.61 & 46.02 & 47.36 & 3.47 & 3.47 & 0.96 \\
				& PointNet2-SSG & - & 56.12 & 46.62 & 46.70 & 46.62 & 58.99 & 1.46 & 1.46 & 2.78 \\
				& PointNet2-MSG & - & 67.63 & \textbf{65.24} & 65.60 & \textbf{65.24} & 68.41 & 1.73 & 1.73 & 8.50 \\
				& SONet & - & 41.01 & 24.26 & 22.41 & 24.26 & 24.79 & 2.66 & 2.66 & 0.99 \\
				& PPFNet & - & 64.03 & 63.96 & 65.35 & 63.96 & 69.75 & 0.28 & 0.28 & 2.37 \\
				& PointCNN & - & 54.68 & 47.23 & 50.42 & 47.23 & 59.79 & 0.27 & 0.27 & 9.84 \\
				& PointWeb & - & 32.37 & 19.16 & 16.13 & 19.16 & 14.59 & 0.78 & 0.78 & 26.12 \\
				& PointConv & - & 50.36 & 47.45 & 43.85 & 47.45 & 44.86 & 19.56 & 19.56 & 7.05 \\
				& RSCNN & - & 48.20 & 30.08 & 25.90 & 30.08 & 30.43 & 1.28 & 1.28 & 1.79 \\
				& PointMLP & - & 48.92 & 40.61 & 41.27 & 40.61 & 48.72 & 13.23 & 13.23 & 13.07 \\
				& PointSCNet & - & 69.06 & 64.27 & \textbf{67.70} & 64.27 & \textbf{73.53} & 1.82 & 1.82 & 5.77 \\
				& RepSurf & - & 67.63 & 59.61 & 59.47 & 59.61 & 59.74 & 1.47 & 1.47 & 4.92 \\
				& PointKAN & - & 48.92 & 41.64 & 43.27 & 41.64 & 56.17 & \textbf{0.16} & 0.16 & \textbf{0.75} \\
				& DELA & - & 32.37 & 17.98 & 11.91 & 17.98 & 9.60 & 5.33 & 5.33 & 1.41 \\
				\midrule
				\multirow{8}{*}{Attention-based} & PCT & - & 46.04 & 36.64 & 36.71 & 36.64 & 37.76 & 2.87 & 2.87 & 3.49 \\
				& P2P & - & 31.65 & 17.79 & 13.86 & 17.79 & 18.11 & 25.04 & 25.04 & 11.08 \\
				& PointTNT & - & 57.55 & 48.74 & 45.45 & 48.74 & 44.70 & 3.93 & 3.93 & 3.15 \\
				& GlobalTransformer & - & 58.27 & 53.45 & 51.84 & 53.45 & 55.14 & 3.75 & 3.75 & 4.01 \\
				& PVT & - & 61.15 & 53.08 & 50.81 & 53.08 & 60.61 & 9.16 & 9.16 & 18.97 \\
				& PointTransformer & - & 59.71 & 56.62 & 52.77 & 56.62 & 51.27 & 3.26 & 3.26 & 2.07 \\
				& PointTransformerV2 & - & 59.71 & 54.75 & 53.54 & 54.75 & 64.53 & 9.59 & 9.59 & 9.41 \\
				& PointTransformerV3 & - & 50.36 & 41.66 & 37.40 & 41.66 & 36.80 & 12.88 & 12.88 & 7.70 \\
				\midrule
				\multirow{7}{*}{Graph-based} & DGCNN & - & 58.99 & 59.35 & 54.11 & 59.35 & 53.99 & 1.80 & 1.80 & 4.63 \\
				& DeepGCN & - & 57.55 & 53.27 & 51.40 & 53.27 & 50.98 & 2.21 & 2.21 & 14.91 \\
				& CurveNet & - & 61.87 & 58.57 & 58.71 & 58.57 & 59.66 & 2.12 & 2.12 & 5.96 \\
				& GDAN & - & 60.43 & 53.55 & 51.90 & 53.55 & 52.52 & 0.93 & 0.93 & 7.40 \\
				& MS-DGCNN & - & 58.99 & 56.15 & 54.33 & 56.15 & 55.89 & 1.55 & 1.55 & 5.44 \\
				& KAN-DGCNN & - & 56.12 & 48.83 & 44.92 & 48.83 & 48.63 & 1.40 & 1.40 & 4.10 \\
				& MS-DGCNN++ & - & 57.55 & 52.61 & 53.28 & 52.61 & 61.64 & 1.81 & 1.81 & 6.91 \\
				\midrule
				\multirow{27}{*}{Self-supervised} & \multirow{6}{*}{PointMAE} & FF & 63.31 & 60.88 & 57.47 & 60.88 & 57.92 & 22.09 & 22.09 & 2.16 \\
				&  & DAPT & 53.24 & 49.70 & 49.30 & 49.70 & 52.89 & 22.89 & 1.06 & 2.04 \\
				&  & IDPT & 64.75 & 61.39 & 59.97 & 61.39 & 60.06 & 23.52 & 1.70 & 1.82 \\
				&  & PPT & 56.83 & 48.21 & 47.78 & 48.21 & 48.56 & 22.78 & 1.04 & 2.77 \\
				&  & GST & 64.75 & 60.18 & 58.99 & 60.18 & 58.71 & 22.45 & 0.62 & 2.82 \\
				&  & VPT & 53.96 & 51.60 & 53.67 & 51.60 & 59.38 & 22.18 & 0.36 & 1.31 \\
				\cmidrule(lr){2-11}
				& \multirow{6}{*}{ACT} & FF & 46.76 & 30.65 & 28.23 & 30.65 & 27.87 & 22.09 & 22.09 & 2.17 \\
				&  & DAPT & 53.96 & 48.36 & 44.47 & 48.36 & 43.72 & 22.89 & 1.06 & 2.06 \\
				&  & IDPT & 57.55 & 47.34 & 47.14 & 47.34 & 58.77 & 23.52 & 1.70 & 1.97 \\
				&  & PPT & 61.87 & 54.35 & 53.17 & 54.35 & 57.20 & 22.78 & 1.04 & 3.00 \\
				&  & GST & \textbf{69.78} & 64.09 & 64.21 & 64.09 & 65.21 & 22.45 & 0.62 & 2.88 \\
				&  & VPT & 56.12 & 49.77 & 49.02 & 49.77 & 60.29 & 22.18 & 0.36 & 1.31 \\
				\cmidrule(lr){2-11}
				& \multirow{6}{*}{RECON} & FF & 39.57 & 26.56 & 26.67 & 26.56 & 38.01 & 43.12 & \textbf{0.01} & 2.18 \\
				&  & DAPT & 53.24 & 49.09 & 48.85 & 49.09 & 51.29 & 22.89 & 1.06 & 1.97 \\
				&  & IDPT & 58.27 & 48.88 & 47.24 & 48.88 & 46.09 & 23.52 & 1.70 & 1.98 \\
				&  & PPT & 57.55 & 58.02 & 56.06 & 58.02 & 56.17 & 22.78 & 1.04 & 2.82 \\
				&  & GST & 67.63 & 60.15 & 63.19 & 60.15 & 69.53 & 22.45 & 0.62 & 2.63 \\
				&  & VPT & 54.68 & 50.93 & 48.86 & 50.93 & 53.25 & 22.18 & 0.36 & 1.33 \\
				\cmidrule(lr){2-11}
				& \multirow{6}{*}{PointGPT} & FF & 61.87 & 52.45 & 52.99 & 52.45 & 54.99 & 29.23 & 29.23 & 2.66 \\
				&  & DAPT & 61.15 & 53.34 & 53.21 & 53.34 & 53.42 & 22.78 & 0.95 & 1.94 \\
				&  & IDPT & 61.87 & 48.40 & 50.13 & 48.40 & 58.41 & 23.49 & 1.70 & 2.10 \\
				&  & PPT & 61.87 & 54.60 & 55.80 & 54.60 & 61.33 & 22.78 & 0.99 & 3.48 \\
				&  & GST & 62.59 & 52.87 & 52.41 & 52.87 & 54.96 & 22.45 & 0.62 & 2.87 \\
				&  & VPT & 50.36 & 37.53 & 35.64 & 37.53 & 36.05 & 22.15 & 0.36 & 1.44 \\
				\cmidrule(lr){2-11}
				& PointM2AE & FF & 61.87 & 57.05 & 53.92 & 57.05 & 52.96 & 12.82 & 12.82 & 5.98 \\
				& PointBERT & FF & 61.87 & 56.02 & 56.37 & 56.02 & 58.71 & 22.06 & 22.06 & 2.11 \\
				& PCP & FF & 46.04 & 38.83 & 33.23 & 38.83 & 35.67 & 22.34 & 22.34 & 2.40 \\
				\bottomrule
			\end{tabular}
		\end{adjustbox}
	\end{table}
	
	\clearpage
	
	
	\begin{table}[htbp]
		\centering
		\scriptsize
		\caption{STPCTLS per-class recall (\%).}
		\label{tab:cv_perclass}
		\begin{adjustbox}{center}
			\begin{tabular}{p{2cm}p{2.5cm}p{0.8cm}p{1.2cm}p{1.2cm}p{1.2cm}p{1.2cm}p{1.2cm}p{1.2cm}p{1.2cm}}
				\toprule
				Category & Model & Strategy & Rec(Buche) (\%) & Rec(Douglasie) (\%) & Rec(Eiche) (\%) & Rec(Esche) (\%) & Rec(Fichte) (\%) & Rec(Kiefer) (\%) & Rec(Roteiche) (\%) \\
				&  &  & $\downarrow$ n=33 & $\downarrow$ n=37 & $\downarrow$ n=4 & $\downarrow$ n=8 & $\downarrow$ n=32 & $\downarrow$ n=5 & $\downarrow$ n=20 \\
				\midrule
				\multirow{15}{*}{Point-based} & PointNet & - & 54.55 & 43.24 & 75.00 & 12.50 & 71.88 & 20.00 & 45.00 \\
				& PointNet2-SSG & - & 78.79 & 59.46 & 50.00 & 37.50 & 65.62 & 20.00 & 15.00 \\
				& PointNet2-MSG & - & 75.76 & 45.95 & 75.00 & 50.00 & 75.00 & 40.00 & \textbf{95.00} \\
				& SONet & - & 60.61 & 64.86 & 0.00 & 0.00 & 34.38 & 0.00 & 10.00 \\
				& PPFNet & - & 81.82 & 70.27 & \textbf{100.00} & 50.00 & 40.62 & 40.00 & 65.00 \\
				& PointCNN & - & 69.70 & 54.05 & 50.00 & 12.50 & 59.38 & 40.00 & 45.00 \\
				& PointWeb & - & 12.12 & 21.62 & 0.00 & 0.00 & 12.50 & 20.00 & 0.00 \\
				& PointWeb & - & 45.45 & 32.43 & 0.00 & 0.00 & 56.25 & 0.00 & 0.00 \\
				& PointConv & - & 54.55 & 43.24 & 75.00 & 12.50 & 71.88 & 40.00 & 35.00 \\
				& RSCNN & - & 93.94 & 21.62 & 0.00 & 0.00 & 75.00 & 0.00 & 20.00 \\
				& PointMLP & - & 60.61 & 40.54 & 25.00 & 0.00 & 53.12 & 40.00 & 65.00 \\
				& PointSCNet & - & \textbf{93.94} & 54.05 & 75.00 & 50.00 & 71.88 & 40.00 & 65.00 \\
				& RepSurf & - & 75.76 & 62.16 & 50.00 & 37.50 & 71.88 & 40.00 & 80.00 \\
				& PointKAN & - & 42.42 & 54.05 & 50.00 & 12.50 & 62.50 & 20.00 & 50.00 \\
				& DELA & - & 39.39 & \textbf{86.49} & 0.00 & 0.00 & 0.00 & 0.00 & 0.00 \\
				\midrule
				\multirow{8}{*}{Attention-based} & PCT & - & 63.64 & 45.95 & 0.00 & 25.00 & 46.88 & 40.00 & 35.00 \\
				& P2P & - & 36.36 & 75.68 & 0.00 & 0.00 & 12.50 & 0.00 & 0.00 \\
				& PointTNT & - & 57.58 & 64.86 & 75.00 & 0.00 & 68.75 & 20.00 & 55.00 \\
				& GlobalTransformer & - & 54.55 & 78.38 & 75.00 & 12.50 & 43.75 & 40.00 & 70.00 \\
				& PVT & - & 78.79 & 62.16 & 75.00 & 12.50 & 53.12 & 20.00 & 70.00 \\
				& PointTransformer & - & 72.73 & 56.76 & 75.00 & 12.50 & 59.38 & 60.00 & 60.00 \\
				& PointTransformerV2 & - & 72.73 & 56.76 & 75.00 & 12.50 & 56.25 & 40.00 & 70.00 \\
				& PointTransformerV3 & - & 90.91 & 43.24 & 75.00 & 0.00 & 37.50 & 0.00 & 45.00 \\
				\midrule
				\multirow{7}{*}{Graph-based} & DGCNN & - & 48.48 & 51.35 & 75.00 & 12.50 & 78.12 & \textbf{80.00} & 70.00 \\
				& DeepGCN & - & 57.58 & 54.05 & 75.00 & 12.50 & 68.75 & 40.00 & 65.00 \\
				& CurveNet & - & 69.70 & 54.05 & 75.00 & 37.50 & 68.75 & 40.00 & 65.00 \\
				& GDAN & - & 72.73 & 70.27 & 75.00 & 0.00 & 46.88 & 40.00 & 70.00 \\
				& MS-DGCNN & - & 51.52 & 62.16 & 75.00 & 12.50 & 71.88 & 60.00 & 60.00 \\
				& KANDGCNN & - & 42.42 & 83.78 & 75.00 & 0.00 & 40.62 & 20.00 & 80.00 \\
				& MS-DGCNN++ & - & 69.70 & 81.08 & 75.00 & 12.50 & 25.00 & 40.00 & 65.00 \\
				\midrule
				\multirow{27}{*}{Self-supervised} & \multirow{6}{*}{PointMAE} & FF & 57.58 & 64.86 & 75.00 & 25.00 & 68.75 & 60.00 & 75.00 \\
				&  & DAPT & 57.58 & 45.95 & 75.00 & 12.50 & 71.88 & 40.00 & 45.00 \\
				&  & IDPT & 57.58 & 70.27 & 75.00 & 25.00 & 71.88 & 60.00 & 70.00 \\
				&  & PPT & 63.64 & 59.46 & 50.00 & 25.00 & 59.38 & 20.00 & 60.00 \\
				&  & GST & 69.70 & 54.05 & 50.00 & 37.50 & 75.00 & 60.00 & 75.00 \\
				&  & VPT & 51.52 & 45.95 & 50.00 & 25.00 & 68.75 & 60.00 & 60.00 \\
				\cmidrule(lr){2-10}
				& \multirow{6}{*}{ACT} & FF & 45.45 & 37.84 & 0.00 & 0.00 & 81.25 & 0.00 & 50.00 \\
				&  & DAPT & 51.52 & 43.24 & 75.00 & 0.00 & 68.75 & 20.00 & 80.00 \\
				&  & IDPT & 57.58 & 59.46 & 50.00 & 12.50 & 71.88 & 20.00 & 60.00 \\
				&  & PPT & 54.55 & 70.27 & 50.00 & 12.50 & 78.12 & 60.00 & 55.00 \\
				&  & GST & 66.67 & 67.57 & 50.00 & \textbf{50.00} & 84.38 & 60.00 & 70.00 \\
				&  & VPT & 54.55 & 51.35 & 75.00 & 12.50 & 75.00 & 20.00 & 60.00 \\
				\cmidrule(lr){2-10}
				& \multirow{6}{*}{RECON} & FF & 45.45 & 64.86 & 25.00 & 0.00 & 40.62 & 0.00 & 10.00 \\
				&  & DAPT & 54.55 & 37.84 & 75.00 & 0.00 & 81.25 & 40.00 & 55.00 \\
				&  & IDPT & 66.67 & 56.76 & 50.00 & 0.00 & 68.75 & 40.00 & 60.00 \\
				&  & PPT & 54.55 & 45.95 & 75.00 & 12.50 & 78.12 & 80.00 & 60.00 \\
				&  & GST & 81.82 & 72.97 & 50.00 & 50.00 & 56.25 & 40.00 & 70.00 \\
				&  & VPT & 54.55 & 43.24 & 75.00 & 12.50 & 81.25 & 40.00 & 50.00 \\
				\cmidrule(lr){2-10}
				& \multirow{6}{*}{PointGPT} & FF & 81.82 & 45.95 & 25.00 & 37.50 & 71.88 & 40.00 & 65.00 \\
				&  & DAPT & 81.82 & 54.05 & 50.00 & 25.00 & 62.50 & 40.00 & 60.00 \\
				&  & IDPT & 66.67 & 62.16 & 25.00 & 25.00 & 75.00 & 20.00 & 65.00 \\
				&  & PPT & 78.79 & 54.05 & 50.00 & 25.00 & 59.38 & 40.00 & 75.00 \\
				&  & GST & 72.73 & 72.97 & 25.00 & 37.50 & 46.88 & 40.00 & 75.00 \\
				&  & VPT & 78.79 & 27.03 & 0.00 & 37.50 & 59.38 & 0.00 & 60.00 \\
				\cmidrule(lr){2-10}
				& PointM2AE & FF & 57.58 & 56.76 & 50.00 & 12.50 & \textbf{87.50} & 80.00 & 55.00 \\
				& PointBERT & FF & 66.67 & 64.86 & 50.00 & 25.00 & 65.62 & 60.00 & 60.00 \\
				& PCP & FF & 45.45 & 35.14 & 0.00 & 0.00 & 81.25 & 80.00 & 30.00 \\
				\bottomrule
			\end{tabular}
		\end{adjustbox}
	\end{table}

\end{document}